\documentclass[conference]{IEEEtran}
\IEEEoverridecommandlockouts
\pdfoutput=1
\usepackage{cite}
\usepackage{amsmath,amssymb,amsfonts}
\usepackage{algorithmic}
\usepackage{graphicx}
\usepackage{textcomp}
\usepackage{xcolor}
\def\BibTeX{{\rm B\kern-.05em{\sc i\kern-.025em b}\kern-.08em
    T\kern-.1667em\lower.7ex\hbox{E}\kern-.125emX}}
\usepackage{booktabs}
\usepackage{cuted}
\usepackage{newfloat}
\usepackage{multirow}
\usepackage{amsmath}
\usepackage{amssymb}

\usepackage{pifont}
\usepackage{arydshln}

\usepackage{array}
\usepackage{float}
\usepackage{cite}
\usepackage{colortbl}
\definecolor{darkred}{rgb}{0.7, 0.1, 0.1}
\definecolor{darkgreen}{rgb}{0.1, 0.6, 0.1}
\definecolor{robo_blue}{RGB}{99,  113,  250}
\definecolor{robo_red}{RGB}{240,  99,  75}
\definecolor{robo_green}{RGB}{0,  180,  139}
\usepackage{adjustbox}
\usepackage{hyperref}
\definecolor{icmeblue}{rgb}{0.21, 0.49, 0.74}

\usepackage{capt-of}

\begin{document}

\title{Create Anything Anywhere: Layout-Controllable Personalized Diffusion Model for Multiple Subjects\\
\thanks{*Corresponding authors}
}
\author{
    Wei Li\textsuperscript{1} \and
    Hebei Li\textsuperscript{1} \and
    Yansong Peng\textsuperscript{1}\and
    Siying Wu\textsuperscript{2,*} \and
    Yueyi Zhang\textsuperscript{1} \and
    Xiaoyan Sun\textsuperscript{1,2,*} \and
\textsuperscript{1}MoE Key Laboratory of Brain-inspired Intelligent Perception and Cognition, \\ University of Science and Technology of China, Hefei, China \and
\textsuperscript{2}Institute of Artificial Intelligence, Hefei Comprehensive National Science Center, Hefei, China\\
{\tt\small \{weili2023, lihebei, pengyansong, wsy315\}@mail.ustc.edu.cn, 
\{zhyuey, sunxiaoyan\}@ustc.edu.cn
}
}



\maketitle

\begin{strip}
\centering
\vspace{-2cm}
\includegraphics[width=\textwidth]{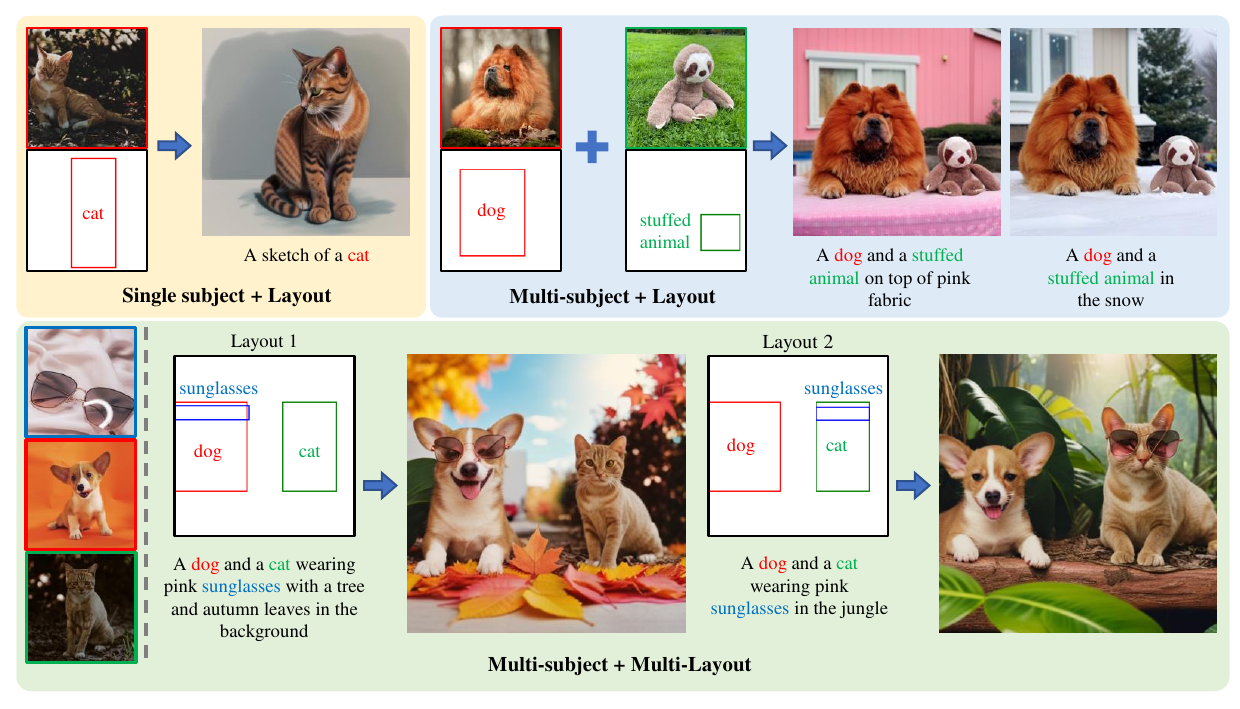}
\vspace{-1cm}

\captionof{figure}{Illustration of LCP-Diffusion, a novel personalized layout-controllable framework. It supports single-subject (top-left), multi-subject (top-right), and multi-layout (bottom) scenarios, accommodating flexible input combinations of multiple reference images, text prompts, and layouts. LCP-Diffusion preserves subject identity, aligns with textual descriptions, and precisely adheres to layout constraints simultaneously, which allows “creating anything anywhere”.}
\vspace{-0.25cm}
\label{fig:teaser}
\end{strip}

\begin{abstract}
Diffusion models have significantly advanced text-to-image generation, laying the foundation for the development of personalized generative frameworks. However, existing methods lack precise layout controllability and overlook the potential of dynamic features of reference subjects in improving fidelity. In this work, we propose Layout-Controllable Personalized Diffusion (LCP-Diffusion) model, a novel framework that integrates subject identity preservation with flexible layout guidance in a tuning-free approach. Our model employs a Dynamic-Static Complementary Visual Refining module to comprehensively capture the intricate details of reference subjects, and introduces a Dual Layout Control mechanism to enforce robust spatial control across both training and inference stages. Extensive experiments validate that LCP-Diffusion excels in both identity preservation and layout controllability. To the best of our knowledge, this is a pioneering work enabling users to “create anything anywhere”.

\end{abstract}


\section{Introduction}
\label{sec:intro}
Recent advances in diffusion models~\cite{rombach2022high, zhang2024trip, 2025motionpro} have revolutionized text-to-image generation, paving the way for personalized generation models~\cite{gal2023an,  ruiz2023dreambooth, kumari2023multi} that aim to accurately reproduce visual concepts from reference images across diverse scenarios, ensuring the generated images are finely tailored to both text prompts and reference images. Personalized generation models provide users with greater autonomy to customize images that align precisely with their preferences,  free from the constraints of random creation and generic styles.

However, personalized generation task remains inherently challenging due to the need for fine-grained control throughout the generation process. Existing methods, both fine-tuned and tuning-free, strive to achieve fine-grained feature extraction that preserves high-fidelity details of the provided reference subjects. Classical fine-tuning methods,  such as Textual Inversion~\cite{gal2023an},  DreamBooth~\cite{ruiz2023dreambooth} and Custom Diffusion~\cite{kumari2023multi} learn a unique embedding concerning specific subject. However, they fall short in capturing intricate subject features and incur high costs for test-time tuning and storage. Tuning-free methods address the resource limitations by eliminating the need for test-time tuning. Diffusion-based tuning-free approaches like SSR-Encoder~\cite{zhang2024ssr} blend subject embeddings into text prompt embeddings to capture enriched cross-modal features. MLLM-based tuning-free approaches,  including BLIP-Diffusion~\cite{li2024blip},  Emu2~\cite{sun2024generative},  and Kosmos-G~\cite{pan2023kosmos} adopt large-scale pre-training for representation learning, excelling in texture modification.
Nevertheless,  all existing methods are confronted with two primary challenges: 1) difficulty in preserving subject identity; 2) lack of layout control. First, these methods struggle with maintaining subject identity, particularly in zero-shot scenarios. They generally overlook variations in subject details,  such as poses and views,  which ultimately limits their overall expressiveness. Second, they lack control over layout, restricting users' ability to specify the subject’s location during image generation. Although InstanceDiffusion~\cite{wang2024instancediffusion} enables instance-level control with layout conditions, it fails to preserve subject identity. An overview of recent state-of-the-art personalized models can be found in Table II of the supplementary material.

In this paper,  we propose a tuning-free \textbf{L}ayout-\textbf{C}ontrollable \textbf{P}ersonalized \textbf{Diffusion} framework (\textbf{LCP-Diffusion}), designed to achieve fine-grained control throughout the personalized generation process in two ways,  1) extracting dynamic features from subject variations and static details from static images in complementary to preserve subject identity; 2) incorporating layout information into the generation process, empowering users to specify the location of the reference target in the generated image. To be specific, first, we design a novel Dynamic-Static Complementary Visual Refining (D-SCVR) module,  which consists of a Dynamic Adaptive Encoder (DA-Encoder) and a Static Detail Refiner. The DA-Encoder novelly extracts dynamic features from subject variations, such as different perspectives, poses in video data as well as varying colors and texture in augmented image data. The static detail refiner enhances the fidelity of static details derived from the reference image. Thus we can capture dynamic-static complementary visual features of the reference subject, and alleviate copy-paste effect.  Second,  we introduce a Dual Layout Control (DLC) Mechanism, which combines a Layout-Aware (LA) module for encoding grounding information and a Box-Constrained Cross-Attention Regulation to direct the model's attention to specified regions during inference.

To sum up,  our contributions are as follows:



\begin{itemize}
    \item  We propose LCP-Diffusion, a novel layout-controllable personalized text-to-image framework, which enables subject identity preservation and precise layout control simultaneously in a tuning-free approach, empowering users to “create anything anywhere”.

    \item We design a D-SCVR module to capture dynamic and static features complementarily, and a DLC mechanism to incorporate accurate layout control. 

    

    
    \item Extensive experiments demonstrate that our method excels in both identity preservation and layout controllability, providing users with enhanced creative freedom. 
\end{itemize}


\section{Related Work}
\noindent\textbf{Personalized Text-to-Image Diffusion Models.} Personalized text-to-image generation, also named subject driven generation or customization, seeks to align with users' own concept generation based on reference subject image and text prompt. These methods can be simply classified into two categories: test-time finetuning method and tuning-free method. Test-time finetuning methods\cite{gal2023an, ruiz2023dreambooth,  kumari2023multi} often involve binding visual concepts with specific identifiers to embed the specific concept into the output domain of the diffusion model. For instance,  DreamBooth\cite{ruiz2023dreambooth} adjusts all parameters of Stable Diffusion~\cite{rombach2022high}. In contrast, Textual Inversion~\cite{gal2023an} focuses on solely optimizing word vectors for new concepts. Custom Diffusion\cite{kumari2023multi} extend similar idea to multi-concept. Tuning-free methods typically design a structure to integrate visual and textual tokens together. IP-Adapter~\cite{ye2023ip} uses a decoupled cross-attention mechanism to utilize image prompt as reference. Kosmos-G~\cite{pan2023kosmos} and $\lambda$-ECLIPSE~\cite{patel2024lambda} share the same idea of bridging the modality gap using CLIP~\cite{radford2021learning} as an anchor. However, all these methods struggle with preserving fine-grained details accounting for subject variations and lack layout controllability. AnyDoor~\cite{chen2023anydoor} can achieve location controllability but it targets at image composition without text prompts, lacking text-to-image generation ability. The concurrent work MS-Diffusion~\cite{wang2024ms}, built on SDXL, also explores layout-controlled customization, but still faces challenges in precise layout control.

\noindent {\textbf{Controllable Text-to-Image Generation.} Controllable diffusion models generate images with specific content via various control conditions, making it applicable in a wide range of fields. ControlNet~\cite{zhang2023adding} leverages a duplicate U-Net encoder structure to guide generation with conditional control. For layout-guided diffusion models,  LayoutDiffusion~\cite{Zheng_2023_CVPR} and GLIGEN~\cite{li2023gligen} incorporate the positions and labels of the bounding boxes into the model to enbale layout understanding. Layout-Guidance~\cite{chen2024training} and BoxDiff~\cite{Xie_2023_ICCV} explore a training-free method by modulating attention maps during the inference phase. However, in the context of personalized generation tasks, layout controllability remains underexplored.

\begin{figure*}[!t]
    \centering
    \includegraphics[width=\textwidth]{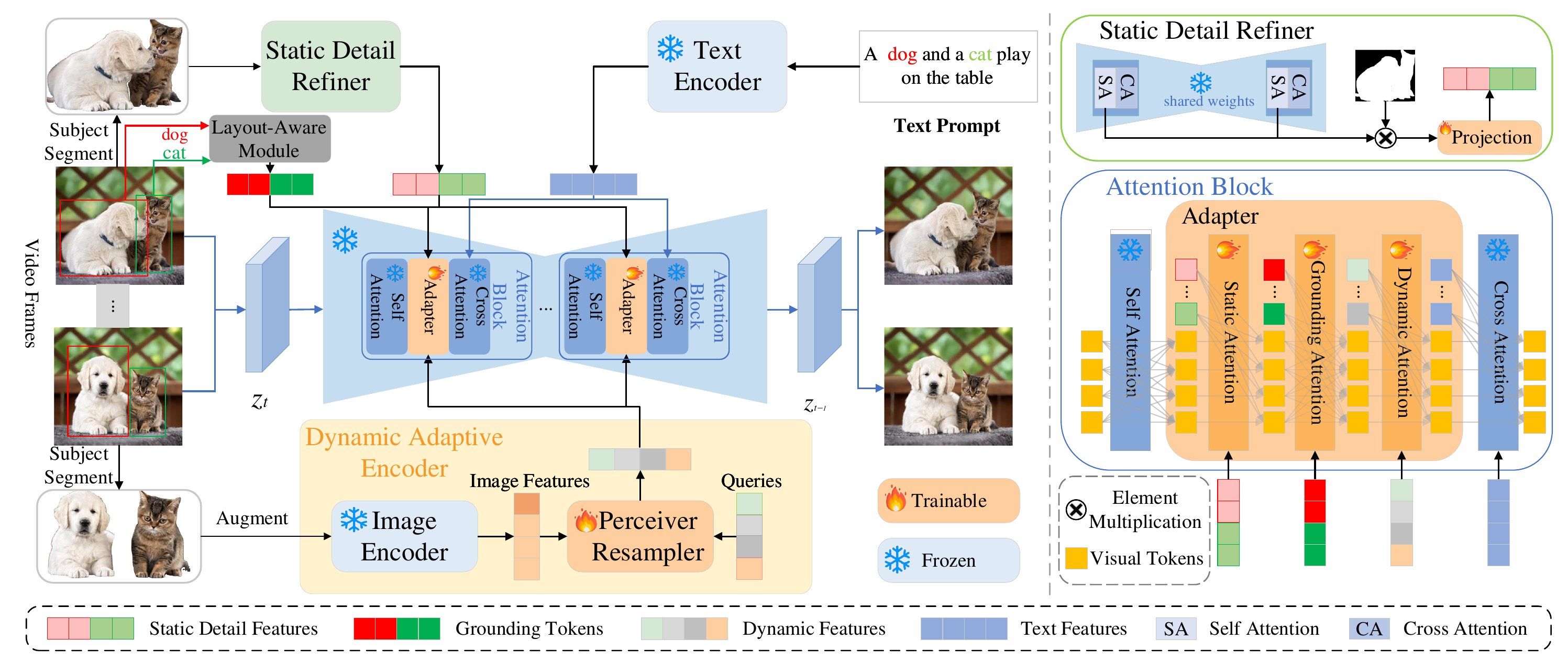}
        \vspace{-0.5cm}
    \caption{Left: Overview of the proposed framework; Right: The structures of Static Detail Refiner and Attention Block in UNet.}
    \label{fig:pipeline}
\end{figure*}
\setlength{\dbltextfloatsep}{1pt}   
\setlength{\textfloatsep}{1pt}       
\setlength{\intextsep}{1pt}          

\section{Proposed Method}\label{sec:method}

\subsection{Preliminaries}
\noindent\textbf{Stable Diffusion.} 
Our method is built on top of Stable Diffusion~\cite{rombach2022high}, which uses UNet as a backbone to denoise from an initial Gaussian noise $z \sim \mathcal{N}(0, \mathbf{I})$ in the latent space. The overall training objective is to minimize the MSE loss, conditioned on additional input $c$, such as text, layout:
\begin{equation}
\label{diffusion_loss}
\mathcal{L} = \mathbb{E}_{z,  c,  t,  \epsilon \sim \mathcal{N}(0, \mathbf{I})} \left[ \left\lVert \epsilon - \epsilon_{\theta}(z_t,  c,  t) \right\rVert_2^2 \right].
\end{equation}
$\epsilon_{\theta}$ represents a U-Net trained to denoise a normally distributed noise $\epsilon$ with the noisy latent $z_t$, current timestep $t$ and conditional input $c$.

\noindent\textbf{Image Prompt Adapter.}
To enable capatibility with image prompt in pre-trained text-to-image diffusion models without fine-tuning the entire U-Net, IP-Adapter~\cite{ye2023ip} introduces a decoupled cross-attention strategy:
\begin{equation}
\boldsymbol{z}^{\textit{new}} = \text{CrossAttn}(z,  c_t) + \lambda \cdot \text{CrossAttn}(z,  c_i), 
\end{equation}
\begin{equation}
\text{CrossAttn}(z,  c_{t/i}) = \text{Softmax}\left(\frac{\mathbf{Q}(\mathbf{K_{t/i}})^\top}{\sqrt{d_k}}\right)\mathbf{V_{t/i}}, 
\end{equation}
where $z$ denotes visual latents in the UNet.  $c_t$ and $c_i$ represent text features and image features respectively. $d_k$ stands for the dimension of key vectors. $\mathbf{K_t}$,$\mathbf{V_t}$ are the  key,  and value matrices from text features, while $\mathbf{K_i}$,$\mathbf{V_i}$ are from image features. Given the query features $z$ and the image features $c_i$, \(\mathbf{Q} = z\mathbf{W}_q\), \(\mathbf{K_i} = c_i \mathbf{W}^i_k\), \(\mathbf{V_i} = c_i \mathbf{W}^i_v\), among which only \(\mathbf{W}^i_k\) and \(\mathbf{W}^i_v\) are trainable. \(\lambda\) serves as a weight coefficient,  typically set to a default value of 1.

\subsection{Dynamic-Static Complementary Visual Refining Module}  
The D-SCVR module integrates a Dynamic Adaptive Encoder to capture dynamic features from subject variations and a Static Detail Refiner to polish static subject details.

\noindent\textbf{Dynamic Adaptive Encoder.} 
We propose Dynamic Adaptive Encoder to capture dynamic features comprehensively from video and augmented image data for detail preservation.
Specifically, for video data, we use the subject content from one frame as the reference subject, while the same subject in another frame serves as a supplement, providing additional detailed features from different poses and views. We use SAM~\cite{kirillov2023segment} to segment subjects in the supplementary frame. For image data, we apply data augmentations such as color jitter, geometric transformations to simulate diverse colors and shapes. The resulting subject segments are then fed into the DA-Encoder,  which comprises a frozen CLIP image encoder and a trainable perceiver resampler~\cite{alayrac2022flamingo}. First,  we extract the unpooled image grid features which consist of 256 visual patch tokens and a CLS token from the penultimate layer of the CLIP image encoder. Next,  we devise a set of $N$ learnable queries to distill subject-relevant information from the image features via the perceiver resampler.
The lightweight resampler is composed of four stacked blocks of perceiver attention combined with feed forward networks (FFN), followed by a projection layer. 
Given a set of learnable visual queries and image grid features, diverse dynamic features can be obtained through several cross-attention layers. Finally, the extracted dynamic features encompassing subject variations are injected into the dynamic attention, detailed in section \ref{sec:adapter}.

\noindent\textbf{Static Detail Refiner.}
To further improve detail fidelity,  we propose to extract self-attention features of the reference subject as complementary static detail features. As shown in Figure \ref{fig:pipeline}, we employ a frozen UNet which shares weights with the UNet backbone in the Stable Diffusion as a static detail refiner. First, we apply a reference subject mask $M_s$ in the form of attention mask to force the model to focus on the subject region. Then the subject image without background is passed through the UNet, conditioned on the category label. Multi-level self-attention maps from the U-Net are aggregated, and the static detail features are distilled through a projection layer. Subsequently, the obtained static detail features $c_{s}$ are fed into static attention in the adapter, detailed in \ref{sec:adapter}. The static detail refiner demonstrates a clear edge over alternatives because it offers multi-level representations at a high resolution and its feature space is naturally aligned with the U-Net backbone in the diffusion model.


\vspace{-0.1cm}
\subsection{Dual Layout Control Mechanism}  
The DLC mechanism incorporates a Layout-Aware module to derive grounding tokens and a Box-Constrained Cross-Attention Regulation to modulate layout during inference.

\noindent\textbf{Layout-Aware Module.}
Our goal is to equip the model with layout awareness while retaining its original capabilities of prompt adherence and identity preservation. We propose a LA module to encode static detail features of reference subjects, corresponding bounding boxes, and entities in the text prompt into integrated grounding tokens through the combination of two multi-layer perceptrons:
\begin{equation}
g = Concat(\text{MLP}\left(c_{s},  \mathcal{F}_{\text{enc}}(b)\right), \text{MLP}\left(c_e,  \mathcal{F}_{\text{enc}}(b)\right)), 
\end{equation}
where $c_e$ refers to the entity embedding encoded by the CLIP text encoder,  while $\mathcal{F}_{\text{enc}}(b)$ denotes the positional representation obtained from bounding box coordinates $b$ through Fourier encoding~\cite{li2023gligen}. $c_{s}$ represents static detail features extracted by the static detail refiner. Subsequently, the grounding tokens $g$ are injected into the grounding attention layer in the adapter.

\noindent\textbf{Box-Constrained Cross-Attention Regulation.}
The proposed box-constrained cross-attention regulation exerts layout guidance by optimizing position and scale loss during inference.
Specifically, in timestep $t$, each cross-attention layer takes intermediate latent feature $z \in \mathbb{R}^{(p \times p)\times d}$, text embedding $c_t \in \mathbb{R}^{n \times d}$ as input and produces the attention map $A^t \in \mathbb{R}^{(p \times p) \times n}$, where $p \in \{64,  32,  16,  8\}$ denotes the spatial dimension of $z$, $n$ is the length of text tokens. For the $k$-th user-specified subject token in the prompt, its attention score with latent pixel at position $(i, j)$ is represented by $A_{(i,j),k}^t$. $B_k$ denotes the bounding box corresponding to the $k$-th user-specified subject token. We design a position constraint loss to enforce the model to focus within the user-specified bounding boxes: 
\begin{equation}
\mathcal{L}_{pos}=\sum_{k=1}^{K} \left( 1 - \frac{\sum_{(i, j) \in B_k} A_{(i, j), k}^t}{\sum_{(i, j)} A_{(i, j), k}^t} \right)^2,
\end{equation}
where $K$ is the number of user-specified subjects.

Moreover, to align the target subject size with the given bounding box, we design a scale constraint loss. First, we transform $B_k$ to a mask $M_k = \mathbf{1}_{\{(i,j) \in B_k\}}$. Next, we create four corner masks, centered at the four vertices of the bounding box, denoted as $V_k=\{V_{k1},V_{k2},V_{k3},V_{k4}\}$. Then we project attention map $A^t_k$,  bounding box mask $M_k$ and corner mask $V_k$ onto the x-axis and y-axis of the attention map respectively and obtain $a^t_{k,x}$,  $a^t_{k,y}$,  $m_{k,x}$,  $m_{k,y}$,  $v_{k,x}$,  $v_{k,y}$ after dimensionality reduction. The scale constraint loss along x-axis is:
\begin{equation}
\label{x-axis loss}
\mathcal{L}_x = \frac{1}{N_x} \sum_{k=1}^{K} \sum_{j=1}^{N_x} v_{k,x}(j) \cdot \left| a^t_{k,x}(j) - m_{k,x}(j) \right|, 
\end{equation}
where $N_x$ denotes normalized width of the bounding box. The scale constraint loss along y-axis $\mathcal{L}_y$ is calculated in the same way as Equation~\ref{x-axis loss}.
The final scale loss can be written as: 
\begin{equation}
\mathcal{L}_{scale} = \mathcal{L}_{x} + \mathcal{L}_{y}.
\end{equation} 

Consequently,  the complete box-constrained loss function guiding the inference process consists of position loss and scale loss. we optimize the combined loss to guide the update direction of the latent vector $z_t$ with a step size of $\alpha_t$:
\begin{equation}
z_t \gets z_t - \alpha_t \cdot \eta \cdot \nabla_{z_t} (\mathcal{L}_{pos} + \mathcal{L}_{scale}), 
\end{equation}
where $\eta $ is a scale factor modulating layout control intensity,  $\alpha_t$ decays linearly at each timestep.

The training-free box-constrained cross-attention regulation can further enhance the layout controllability and effectively alleviates issues of subject omission and inter-subject conflicts within the multi-subject generation framework.

\subsection{Adapter}
We employ a lightweight adapter-based fine-tuning strategy to embed both detailed features and spatial cues into the diffusion model. It consists of static attention layer, grounding attention layer and dynamic attention layer.
\label{sec:adapter} 

\noindent\textbf{Static Attention.}
We inject static detail features $c_s$ extracted from the static detail refiner into the diffusion model through the cross-attention layers to preserve detail fidelity. Inspired by ~\cite{song2024moma},  the static attention output $z_s$ can be formulated as :
\begin{equation}
z_s =  \text{SelfAttn}(z) + \mu \cdot \text{CrossAttn}(z,  c_{s}, M_s) \cdot \alpha, 
\end{equation}
where $\mu$ is a learnable parameter, $\alpha$ is the injection strength, $M_s$ is the reference subject mask. The output $z_s$ is fed into the next grounding attention layer.

\noindent\textbf{Grounding Attention.}
Grounding attention is devised as a gated self-attention. Taking grounding tokens $g$ and static attention output $z_s$ as input, the approach can be formulated in a residual manner based on ${z}_s$:
\begin{equation}
\label{eq:grounding}
z_g = z_s + \beta \cdot \tanh(\gamma) \cdot \text{SelfAttn}\left([z_s,  g]\right), 
\end{equation}
where $\beta$ is a coefficient that adjusts the conditioning strength,  and $\gamma$ is a learnable scalar initially set to 0. After the self-attention operation,  we discard the concatenated grounding embeddings,  retaining only visual queries. Thus we integrate spatial conditions into the visual queries without changing the sequence length. Consequently, we can gradually adapt the model to spatial condition compatibility through a lightweight grounding attention without finetuning the whole model.

\noindent\textbf{Dynamic Attention.}
We follow IP-Adapter~\cite{ye2023ip} to feed the output features as dynamic subject features into trainable decoupled cross-attention layers for image prompt. Combined with the frozen cross-attention for text,  the final output of the decoupled cross-attention $\boldsymbol{z}^{\textit{new}}$ can be formulated as:
\begin{equation}
\boldsymbol{z}^{\textit{new}}= \text{CrossAttn}(z_g,  c_t) + \lambda \cdot \text{CrossAttn}(z_g,  c_d), 
\end{equation}

\noindent where $c_t$ and $c_d$ represent text features and dynamic subject features respectively. 

\vspace{-0.1cm}
\subsection{Training and Inference}
During training,  we optimize the adapter containing three types of attention layers,  perciever resampler,  and MLP in LA module while keeping the pre-trained UNet frozen. 
The training process is conditioned on text features $c_t$, reference subject features $c_s$, $c_d$ and grounding tokens $g$ following the original denoising objective of Stable Diffusion in Equation~\ref{diffusion_loss}.

During inference,  we input reference subject images, corresponding subject masks generated by SAM~\cite{kirillov2023segment} and bounding boxes, text prompts to generate an image where the reference subjects are placed in specified positions. We can also generate without bounding box input, during which we deactivate box-constrained cross-attention regulation.

\begin{figure}[!t]
    \centering
    \vspace{-0.3cm}
    \includegraphics[width=0.49\textwidth]{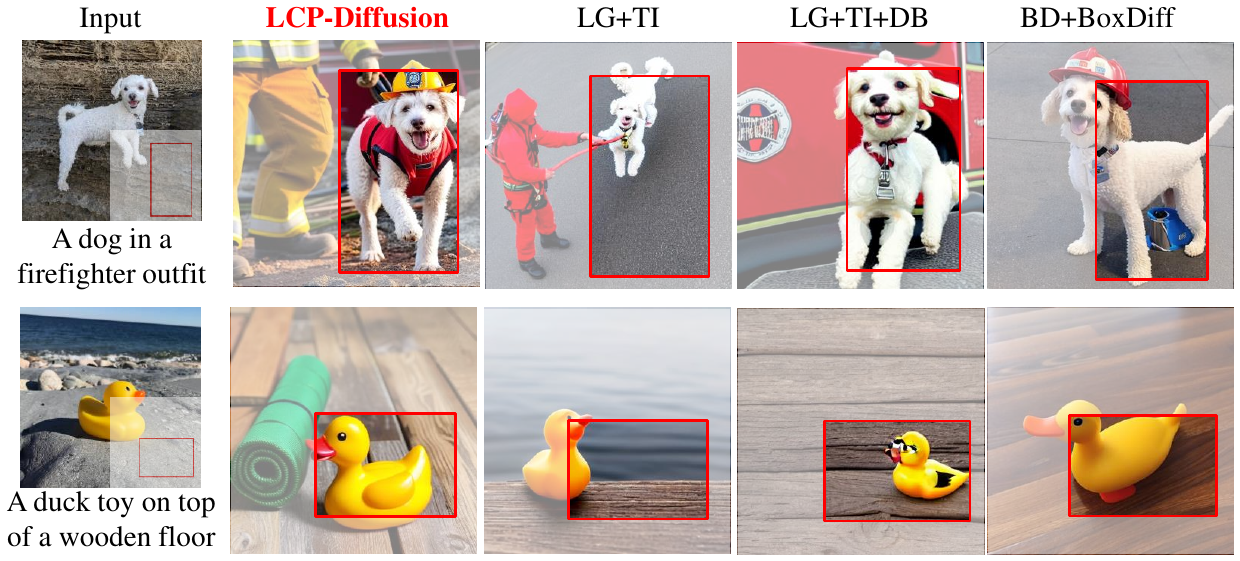}
    \vspace{-0.7cm}
    \caption{Qualitative results show that LCP-Diffusion performs best both in accurate layout control and faithful text prompt alignment.}
    \label{fig:layout_cmp}
\end{figure}

\begin{figure}[!t]
    \centering
    \vspace{-0.4cm}
    \includegraphics[width=0.5\textwidth]{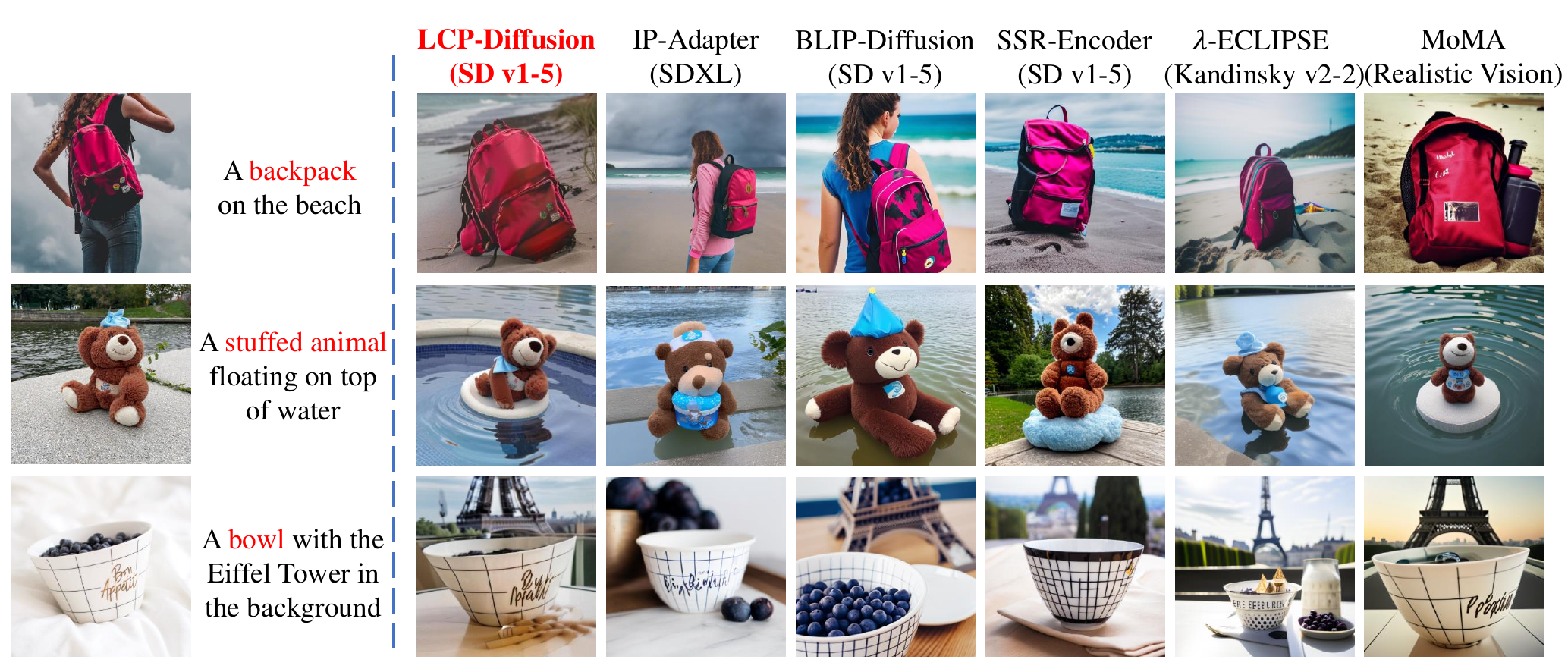}
    \vspace{-0.7cm}
    \caption{Qualitative results show the fine-grained detail preservation ability beyond layout controllability.}
    \label{fig:single_cmp}
\end{figure}

\begin{table}[!t]
\centering
\setlength{\tabcolsep}{2pt} 
\renewcommand{\arraystretch}{1.1} 
\vspace{-0.27cm}
\caption{Quantitative comparison of layout controllable personalized text-to-image methods on DreamBench and MultiBench.}
\vspace{-0.2cm}
\resizebox{\columnwidth}{!}{ 
\begin{tabular}{>{\centering\arraybackslash}m{2.5cm}lcccc}
\toprule
Category & Method & CLIP-I & DINO & CLIP-T & AP/AP\textsubscript{50}/AP\textsubscript{75} \\
\midrule
\multirow{4}{*}{\textit{Single-Subject}} 
& BD + BoxDiff & 0.715 & 0.584 & 0.301 & 10.5/16.9/11.1 \\
& LG + TI & 0.752 & 0.477 & 0.273 & 13.6/27.7/18.5 \\
& LG + TI + DB & 0.775 & 0.598 & 0.307 & 19.7/32.6/21.1 \\
& \textbf{LCP-Diffusion} & \textbf{0.789} & \textbf{0.662} & \textbf{0.330} & \textbf{42.2/64.5/48.7} \\
\midrule
\multirow{1}{*}{\textit{Multi-Subject}} 
& \textbf{LCP-Diffusion} & \textbf{0.717} & \textbf{0.413} & \textbf{0.364} & \textbf{34.6/57.1/39.3} \\
\bottomrule
\end{tabular}
}
\label{tab:tab1}
\end{table}

\section{Experiment and Analysis}\label{sec:exp}  
\vspace{-0.3cm}
\begin{table}[!t] 
    \centering
    \vspace{-0.4cm}
    \caption{Quantitative comparison on LCP-Diffusion and comparing methods for single-subject. * denotes the model is fine-tuned on DreamBench.}
    \vspace{-0.1cm}
    \setlength{\tabcolsep}{10pt} 
    \renewcommand{\arraystretch}{1} 
    \begin{tabular}{lccc}
    \toprule
    Method & CLIP-I  & DINO & CLIP-T \\ 
    \midrule
    \rowcolor{robo_red!20}\multicolumn{4}{l}{\textcolor{robo_red}{\textit{Fine-Tuning Methods}}} \\
    Textual Inversion~\cite{gal2023an} & 0.780 & 0.569 & 0.255 \\
    DreamBooth~\cite{ruiz2023dreambooth} & 0.803 & 0.668 & 0.305 \\
    Custom-Diffusion~\cite{kumari2023multi} & 0.790 & 0.643 & 0.305 \\
    BLIP-Diffusion\textsuperscript{*}~\cite{gal2023an} & \textbf{0.805} & 0.670 & 0.302 \\
    $\lambda$-ECLIPSE\textsuperscript{*}~\cite{gal2023an} & 0.796 & 0.682 & 0.304 \\
    \textbf{LCP-Diffusion\textsuperscript{*}} & 0.803 & \textbf{0.691} & \textbf{0.319} \\
    \midrule
    \rowcolor{robo_blue!20}\multicolumn{4}{l}{\textcolor{robo_blue}{\textit{Tuning-Free Methods}}} \\
    GLIGEN~\cite{li2023gligen} & 0.723 & 0.501 & 0.305 \\
    BLIP-Diffusion~\cite{li2024blip} & 0.779 & 0.594 & 0.300 \\
    IP-Adapter~\cite{ye2023ip} & 0.810 & 0.613 & 0.292 \\
    SSR-Encoder~\cite{zhang2024ssr} & \textbf{0.821} & 0.612 & 0.308 \\
    Emu2~\cite{sun2024generative} & 0.765 & 0.563 & 0.273 \\
    $\lambda$-ECLIPSE~\cite{patel2024lambda} & 0.783 & 0.613 & 0.307 \\
    MoMA~\cite{song2024moma} & 0.801 & 0.618 & 0.324 \\
    \textbf{LCP-Diffusion} & 0.789 & \textbf{0.662} & \textbf{0.330} \\
    \bottomrule
    \end{tabular}

    \label{tab:tab2}
\end{table}

In this section,  we evaluate our LCP-Diffusion from two aspects: layout controllability and identity preservation. We validate the superiority of the proposed LCP-Diffusion through a series of quantitative and qualitative experiments.

\subsection{Experiment Setup}
\textbf{Training data.} 
We collect 86k video clips from open-sourced websites like Pexels.  Furthermore,  we also collect a image subset from COCO Stuff~\cite{caesar2018coco} and OpenImage-V6~\cite{kuznetsova2020open} to expand the diversity of subject categories.

\textbf{Evaluation metrics.}
We evaluate our method comprehensively both in identity preservation ability and layout controllability on DreamBench~\cite{ruiz2023dreambooth} and MultiBench~\cite{patel2024lambda}. Specifically, we compute DINO and CLIP-I to assess subject fidelity, and CLIP-T for text fidelity. To evaluate the location of generated subjects, we apply average precision (AP) by comparing the detected bounding boxes, obtained using the detection model LW-DETR~\cite{chen2024lw}, with the annotated ground truth.

\subsection{Implementation Details} 
We apply Stable Diffusion v1.5~\cite{rombach2022high} as foundation diffusion model and pre-trained CLIP~\cite{radford2021learning} as text and image encoder. The parameters of dynamic attention are initialized from the checkpoint of IP-Adapter~\cite{ye2023ip}, and the parameters of static attention are zero-initialized.
All of the input images are resized to the resolution of $512\times512$. We use the AdamW optimizer with a constant learning rate of 5e-5. We train our model on 4×A800 GPUs with a batch size of 16 per GPU for 100k steps. Inference is performed using DDIM as the sampler,  with a step size of 50 and a guidance scale set to 7.5. The number of learnable query vectors $N$ is set to 16. During training,  the sampling portion of the video data is 70\% while that of image data is 30\%. For video data, we sample two frames from a video clip in random to form a frame pair. 

\subsection{Quantitative Comparison}
Due to the lack of prior work on layout-controlled personalization with diffusion models,  we naively combine existing personalization method with layout guidance techniques as compared methods to evaluate layout controllability,  such as BLIP-Diffusion~\cite{li2024blip} with BoxDiff~\cite{Xie_2023_ICCV},  Dreambooth~\cite{ruiz2023dreambooth} and Textual Inversion~\cite{gal2023an} with backward layout-guidance~\cite{chen2024training}. From Table \ref{tab:tab1}, we observe that our method significantly outperforms others across all metrics.

Table \ref{tab:tab2} shows our quantitative analysis for identity preservation in both zero-shot and finetuning scenarios. LCP-Diffusion achieves the highest scores in DINO and CLIP-T, with a clear lead in CLIP-T. CLIP-I metric tends to favor broader image-text alignment like overall background rather than precise detail matching. In the qualitative comparison below,  we can justify that LCP-Diffusion prioritizes the text prompt interpretation and subject detail fidelity,  which may not be fully reflected in CLIP-I. Further results on multi-subject are shown in Table III of the supplementary material.

\subsection{Qualitative Comparison}
As depicted in Figure \ref{fig:layout_cmp},  LCP-Diffusion excels in accurate layout control while faithfully preserving details from the reference image and aligning with the textual prompt.
Figure \ref{fig:single_cmp} highlights our method’s detail preservation ability on DreamBench. 
Overall,  LCP-Diffusion consistently outperforms other state-of-the-art methods, avoiding common problems of inconsistencies and unnatural interactions. A user study in Section G of the supplementary material supports our findings.

\begin{figure}[t!]
    \centering
    \includegraphics[width=\columnwidth]{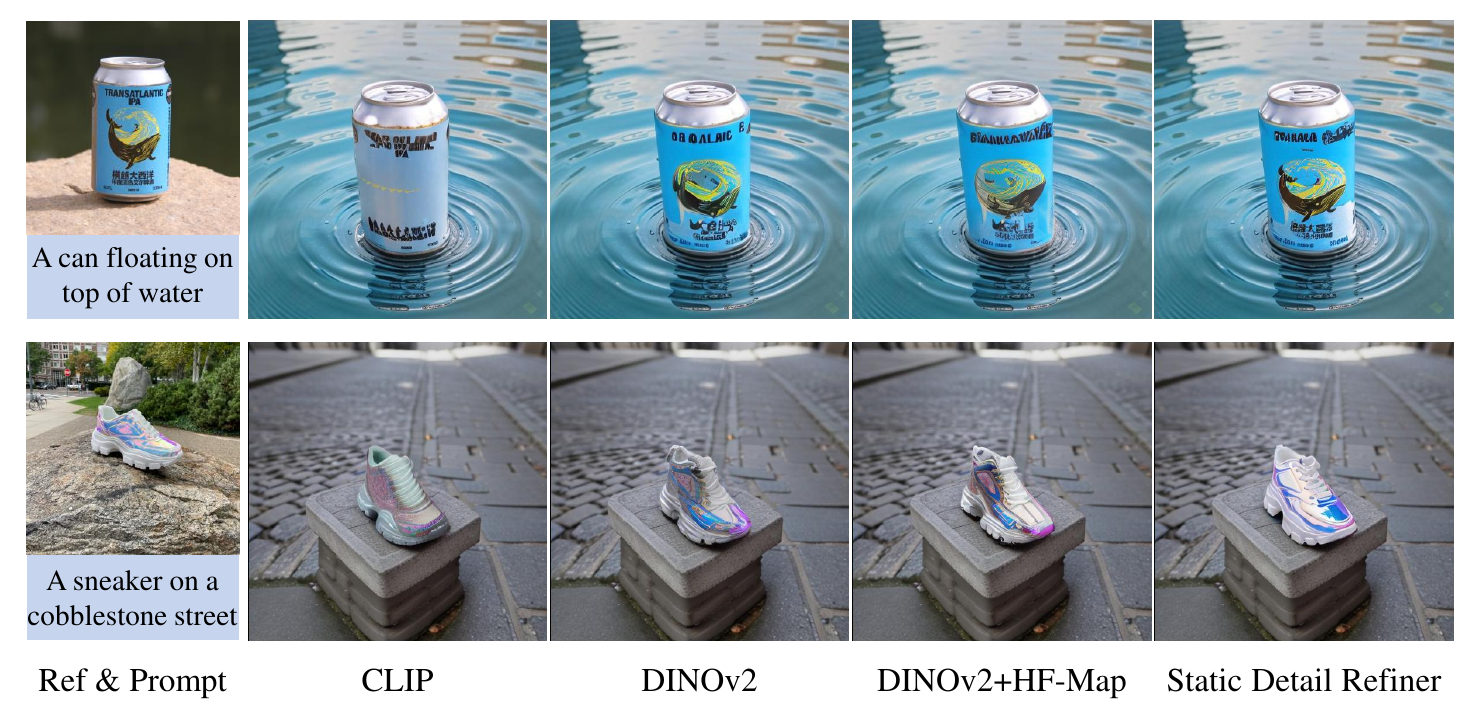}
    \vspace{-0.8cm}
    \caption{An illustration of the superior performance of the designed static detail refiner. }
    \label{fig:ablation1}
\end{figure}

\vspace{-0.3cm}
\subsection{Ablation Studies and Analysis}  

As shown in Table \ref{tab:tab3}, we comprehensively analyze the following components in our method.
(1) augmentation on static images
(2) static detail refiner
(3) layout-aware module 
(4) box-constrained cross-attention regulation.
\begin{table}[!t]
    \centering
    \vspace{-0.4cm}
    \caption{Ablation results on the overall framework.}
    \vspace{-0.2cm}
    \begin{tabular}{lccc}
    \toprule
     & CLIP-I  & DINO & CLIP-T \\
    \midrule
    Baseline & 0.737 & 0.608 & 0.286\\
    + Augmentation & 0.761 & 0.622 &  0.285 \\
    ++ Static Detail Refiner & 0.776 & 0.651 & 0.303 \\
    +++ Grounding Attn & \textbf{0.791} & 0.660 & 0.313\\
    ++++ Cross-Attn Regulation & 0.789 & \textbf{0.662} & \textbf{0.330} \\
    \bottomrule
    \end{tabular}
    \label{tab:tab3}
    
\end{table}
Augmentations such as color jitter and geometric transformations simulate deformations and texture variations, enabling the model to adapt to texture change prompts. Visual comparisons in Figure \ref{fig:ablation1} and quantitative results in the supplementary material validate that the static detail refiner outperforms alternatives such as CLIP,  DINOv2 encoder~\cite{oquab2023dinov2} and HF-Map extractor~\cite{chen2023anydoor}  as a more effective feature extractor. 
Moreover,  box-constrained cross-attention regulation further enhances layout guidance. Although showing a slight decrease in CLIP-I, it demonstrates a notable improvement in CLIP-T. We assume its potential lies in spatially disentangling the cross-attention maps of different semantic elements, thereby mitigating the identity blending problem and improving layout precision during generation.
\vspace{-0.3cm}

\section{Conclusion}\label{sec:conclusion} 
\vspace{-0.2cm}
In this paper, we present a layout-controllable framework to achieve finer-grained control for personalized text-to-image generation. Our design focuses on two key aspects: preserving subject details and enhancing layout controllability. The proposed Dynamic-Static Complementary Visual Refining module effectively captures dynamic and static details from videos and images in a complementary manner, while the Dual Layout Control Mechanism endows the model with robust layout controllability across training and inference stages. The zero-shot framework, allowing users to “create anything anywhere”, is also adaptable to other custom diffusion models, manifesting the potential to extend to broader applications. 

\vspace{-0.2cm}
\section{Acknowledgment}
\vspace{-0.1cm}
This work was in part supported by the National Natural Science Foundation of China under grants
62032006 and 62021001.

\vspace{-0.15cm}
\bibliographystyle{IEEEbib}
\vspace{-0.2cm}
\bibliography{reference}


\end{document}